\documentclass[letterpaper, 10 pt, conference]{ieeeconf}  % Comment this line out if you need a4paper

 \IEEEoverridecommandlockouts                              % This command is only needed if 
\usepackage{amsmath} % assumes amsmath package installed
\usepackage{amssymb}  % assumes amsmath package installed

\usepackage{bm}
\usepackage{srcltx}
\usepackage{cite}    
\usepackage{verbatim}
\usepackage{float}
\usepackage{acronym}
\usepackage{color}
\usepackage[lined,ruled,vlined]{algorithm2e}
\usepackage{graphicx}

\usepackage{paralist}
\usepackage[left=20mm, right=20mm, top=20mm,
  bottom=20mm]{geometry}

\usepackage{tikz}
\usetikzlibrary{shapes,snakes}

\usepackage{placeins}
\usetikzlibrary{arrows,fit,shapes,automata}
\usetikzlibrary{positioning,fit,calc,shapes}
\usetikzlibrary{decorations.fractals}
\usetikzlibrary{decorations.markings}

\usepackage{caption}
\usepackage{subcaption}

 \usepackage{acronym}
\providecommand{\abs}[1]{\left|#1\right|}

\newcommand{\calF}{\mathcal{F}}
\newcommand{\calM}{\mathcal{M}}
\newcommand{\calD}{\mathcal{D}}
\newcommand{\bbE}{\mathbb{E}}
\usepackage{setspace}

\newcommand{\calW}{\mathcal{W}}

\newcommand{\calA}{\mathcal{A}}
\newcommand{\calAP}{\mathcal{AP}}
\newcommand{\AEC}{\mathsf{AEC}}
\newcommand{\truev}{\mathsf{true}}
\newcommand{\falsev}{\mathsf{false}}

    \renewcommand{\Pr}{\mathrm{Pr}}   

 \newtheorem{definition}{Definition}
 \newtheorem{example}{Example}
\newtheorem{problem}{Problem}

 \newtheorem{proposition}{Proposition}

\acrodef{dra}[DRA]{deterministic Rabin automaton}
\acrodef{molp}[MOLP]{multi-objective linear programming}
\acrodef{solp}[SOLP]{single-objective linear programming}

\acrodef{ltl}[LTL]{linear temporal logic}
 \acrodef{mdp}[MDP]{Markov decision process} 
\acrodef{rl}[RL]{reinforcement learning}
\acrodef{momdp}[multi-objective MDP]{multi-objective Markov decision
  process}

\newcommand{\supp}{\mathsf{Support}}
\newcommand{\cost}{\mathsf{Cost}}

\newcommand{\att}{C} 
\newcommand{\act}{\Gamma}

\acrodef{aec}[AEC]{accepting end component}
\usepackage{graphicx}
\graphicspath{{./figure/}}

\title{\LARGE \bf Pareto efficiency in synthesizing shared autonomy policies with temporal logic constraints}

\author{Jie Fu and Ufuk Topcu$^{1}$% <-this % stops a space
  \thanks{This work is supported by AFOSR grant \# FA9550-12-1-0302,
    ONR grant \# N000141310778, and NSF CNS award \#
    1446479.}% <-this % stops a space
  \thanks{$^{1}$Jie Fu and Ufuk Topcu are with the Department of
    Electrical and Systems Engineering, University of Pennsylvania,
    Philadelphia, PA, 19104, USA {\tt\small jief,
      utopcu@seas.upenn.edu}}%
}

\begin{document}

\maketitle
\thispagestyle{empty}
\pagestyle{empty}

%%%%%%%%%%%%%%%%%%%%%%%%%%%%%%%%%%%%%%%%%%%%%%%%%%%%%%%%%%%%%%%%%%%%%%%%%%%%%%%%
\begin{abstract}
  In systems in which control authority is shared by an autonomous
  controller and a human operator, it is important to find solutions
  that achieve a desirable system performance with a reasonable
  workload for the human operator.  We formulate a shared autonomy
  system capable of capturing the interaction and switching control
  between an autonomous controller and a human operator, as well as
  the evolution of the operator's cognitive state during control
  execution. To trade-off human's effort and the performance level,
  e.g., measured by the probability of satisfying the underlying
  temporal logic specification, a two-stage policy synthesis algorithm
  is proposed for generating Pareto efficient coordination and control
  policies with respect to user specified weights. We integrate the
  Tchebychev scalarization method for multi-objective optimization
  methods to obtain a better coverage of the set of Pareto efficient
  solutions than linear scalarization methods. %  In this way, the system
  % designer has the flexibility to construct shared autonomy policies
  % under temporal logic constraints with respect to his/her own
  % preference.
\end{abstract}

\section{Introduction}
%%%%%%%%%%%%%%%%%%%%%%%%%%%%%%%%%%%%%%%%%%%%%%%%%%%%%%%%%%%%%%%%%%%%%%%%%%%%%%%%

Despite the rapid progress in designing fully autonomous systems, many
systems still require human's expertise to handle tasks which
autonomous controllers cannot handle or which they have poor
performance.  Therefore, shared autonomy systems have been developed
to bridge the gap between fully autonomous and fully human operated
systems.  In this paper, we examine a class of shared autonomy
systems, featured by switching control between a human operator and an
autonomous controller to collectively achieve a given control objective.
Examples of such shared autonomy systems include robotic mobile
manipulation \cite{Pitzer2011}, remote tele-operated mobile robots
\cite{kinugawa2001shared}, human-in-the-loop autonomous driving
vehicle \cite{Gnatzig2012,Li2014}.  In particular, we consider
control under temporal logic specifications.

% what problem we need to address. 
% For a system in this class, its performance depends on when the
% control is switched from one entity to another, timeframes for being
% controlled under which entity, as well as other factors that
% potentially influence the performance of the human operator, such as,
% fatigue, boredom, etc.
One major challenge for designing shared autonomy policies under
temporal logic specifications is making trade-offs between two possibly
competing objectives: Achieving the optimal performance for satisfying
temporal logic constraints and minimizing human's effort. Moreover,
human's cognition is an inseparable factor in synthesizing shared
autonomy systems since it directly influences human's performance, for
example, a human may have limited time span of attention and possible
delays in response to a request. Although finding an accurate model of
human cognition is an ongoing challenging topic within cognitive
science, Markov models have been proposed to model and predict human
behaviors in various decision making tasks
\cite{Pentland1999,rothkopf2013modular,mcghan2012human}. Adopting this
modeling paradigm for human's cognition, we propose a formalism for
shared autonomy systems capturing three important components: The
operator, the autonomous controller and the cognitive model of the
human operator, into a stochastic \emph{shared-autonomy
  system}. Precisely, the three components includes a Markov model
representing the fully-autonomous system, a Markov model for the fully
human-operated system, and a Markov model representing the evolution
of human's cognitive states under requests from autonomous controller
to human, or other external events.  The uncertainty in the composed
system comes from the stochastic nature of the underlying dynamical
system and its environment as well as the inherent uncertainty in the
operator's cognition.  Switching from the autonomous controller to the
operator can occur only at a particular set of human's cognitive
states, influenced by requests from the autonomous controller to the
operator, such as, pay more attention, be prepared for a possible
future control action.

Under this mathematical formulation, we transform the problem of
synthesizing a shared autonomy policy that coordinates the operator
and the autonomous controller into solving a \ac{momdp} with temporal
logic constraints: One objective is to optimize the probability of
satisfying the given temporal logic formula, and another objective is
to minimize the human's effort over an infinite horizon, measured by a
given cost function.  The trade-off between multiple objectives is
then made through computing the Pareto optimal set. Given a policy in
this set, there is no other policy that can make it better for one
objective than this policy without making it worse for another
objective. In literature, Pareto optimal policies for \ac{momdp}s have
been studied for the cases of long-run discounted and average rewards
\cite{chatterjee2006markov,chatterjee2007}. The authors in
\cite{FKP12} proposed the weighted-sum method for \ac{momdp}s with
multiple temporal logic constraints by solving Pareto optimal policies
for undiscounted time-bounded reachability or accumulated
rewards. These aforementioned methods are not directly applicable in
our problem due to the time unboundness in both satisfying these
temporal logic constraints and the accumulated cost/reward. To this
end, we develop a novel two-stage optimization method to handle the
multiple objectives and adopt the so-called \emph{Tchebychev
  scalarization method} \cite{perny2010finding} for finding a uniform
coverage of all Pareto optimal points in the policy space, which
cannot be computed via weighted-sum (linear scalarization) methods
\cite{das1997closer} as the latter only allows Pareto optimal
solutions to be found amongst the convex area of the Pareto
front. Finally, we conclude the paper with an algorithm that generates
a Pareto-optimal policy achieving the desired trade-off from
user-defined weights for coordinating the switching control between an
operator and an autonomous controller for a stochastic system with
temporal logic
constraints. % Possible future extensions are discussed.

\section{Preliminaries}
We provide necessary background for presenting the results in this
paper.

A vector in $\mathbb{R}^n$ is denoted $\vec{v}=(v_1,v_2,\ldots, v_n)$
where $v_i, 1\le i\le n$ are the components of $\vec{v}$.  We denote
the set of probability distributions on a set $S$ by $\calD(S)$. Given
a probability distribution $D:S\rightarrow [0,1]$, let $\supp(D(s)) =
\{s\in S\mid D(s)\ne 0\}$ be the set of elements with non-zero
probabilities in $D$.

\subsection{Markov decision processes and control policies}
\begin{definition}
\label{def:labeledmdp}
A \emph{labeled \ac{mdp}} is a tuple $ M= \langle S, \Sigma, D_0,T,
\calAP, L, r, \gamma \rangle $ where $S$ and $\Sigma$ are finite state
and action sets.  $D_0: S\rightarrow \mathbb{R}$ is the initial
probability distribution over states. The transition probability
function $T: S \times \Sigma \times S \rightarrow [0,1]$ is defined
such that given a state $s\in S$ and an action $\sigma\in \Sigma$,
$T(s,\sigma,s')$ gives the probability of reaching the next state
$s'$. $\calAP$ is a finite set of atomic propositions and $L: S
\rightarrow 2^{\calAP}$ is a labeling function which assigns to each
state $s \in S$ a set of atomic propositions $L(s)\subseteq \calAP$
that are valid at the state $s$. $r: S \times \Sigma\times
S\rightarrow \mathbb{R}$ is a reward function giving the immediate
reward $r(s,a,s')$ for reaching the state $s'$ after taking action $a$
at the state $s$ and $\gamma\in (0,1)$ is the reward discount factor. $\blacksquare$
\end{definition}
In this context, $T(s,a)$ gives a probability distribution over the
set of states.  $T(s,a)(s')$ and $T(s,a,s')$ both express the
transition probability from state $s$ to state $s'$ under action $a$
in $M$.  A \emph{path} is an infinite sequence $s_0s_1\ldots$ of
states such that for all $i\ge 0$, there exists $a\in \Sigma$,
$T(s_i,a,s_j)\ne 0$.  We denote $\act(s) \subseteq \Sigma$ to be a set
of actions enabled at the state $s$. That is, for each $a\in \act(s)$,
$\supp(T(s,a))\ne\emptyset$.
% The \emph{structure} of the labeled \ac{mdp} $M$ is the underlying
% graph $\langle S, \Sigma, E\rangle$ where $E \subseteq S\times
% \Sigma\times S$ is the set of labeled edges. $(s,a,s')\in E$ if and
% only if $T(s,a,s')\ne 0 $.

A \emph{randomized policy} in $M$ is a function $f: S^\ast\rightarrow
\mathcal{D}(\Sigma)$ that maps a finite path into a probability
distribution over actions. A deterministic policy is a special case of
randomized policies that maps a path into a single action.  Given a
policy $f$, for a measurable function $\phi$ that maps paths into
reals, we write $\bbE_s^f[\phi]$ (resp. $\bbE_{D_0}^f[\phi]$) for the
expected value of $\phi$ when the \ac{mdp} starts in state $s$
(resp. an initial distribution of states $D_0$) and the policy $f$ is
used. A policy $f$ induces a probability distribution over paths in
$M$. The state reached at step $t$ is a random variable $X_t$ and the
action being taken at state $X_t$ is also a random variable, denoted
$A_t$.

\subsection{Synthesis for \ac{mdp}s with temporal logic constraints}
\label{sub:mdpltl}
We use \ac{ltl} \cite{emerson1990temporal} to specify a set of desired
system properties such as safety, liveness, persistence and
stability. In the following, we present some basic preliminaries for
\ac{ltl} specifications and introduce a product operation for
synthesizing policies in \ac{mdp}s under \ac{ltl} constraints.

A formula in \ac{ltl} is built from a finite set of atomic
propositions $\calAP$, $\truev$, $\falsev$ and the Boolean and
temporal connectives $\land, \lor, \neg, \Rightarrow, \Leftrightarrow$
and $\square$ (always), $\mathcal{U}$ (until), $\lozenge$
(eventually), $\bigcirc$ (next). Given an \ac{ltl} formula $\varphi$
as the system specification, one can always represent it by a \ac{dra}
$\calA_\varphi=\langle Q,2^{\calAP}, \delta,I, \mathsf{Acc} \rangle$
where $Q$ is a finite state set, $2^{\calAP}$ is the alphabet, $I\in
Q$ is the initial state, and $\delta: Q\times 2^{\calAP} \rightarrow
Q$ is the transition function.  The acceptance condition $\mathsf{Acc}$
is a set of tuples $\{(J_i, K_i) \in 2^Q \times 2^Q \mid i =0,
1,\ldots,m\}$. The run for an infinite word $w= w[0]w[1]\ldots \in
(2^{\calAP})^\omega$ is an infinite sequence of states $q_0q_1\ldots
\in Q^\omega$ where $q_0=I$ and $q_{i+1}=\delta(q_i, w[i])$. A run
$\rho=q_0q_1\ldots $ is accepted in $\calA_\varphi$ if there exists at
least one pair $(J_i,K_i) \in \mathsf{Acc}$ such that
$\mathsf{Inf}(\rho)\cap J_i =\emptyset$ and $\mathsf{Inf}(\rho)\cap
K_i \ne \emptyset$ where $\mathsf{Inf}(\rho)$ is the set of states
that appear infinitely often in $\rho$.

We define a product operation between a labeled \ac{mdp} and a
\ac{dra}.

\begin{definition}\label{def:product}
  Given a labeled \ac{mdp} $M= \langle S, \Sigma, D_0,T, \calAP, L, r,
  \gamma \rangle $ and the \ac{dra} $\mathcal{A}_\varphi= \langle Q
  ,2^{\calAP}, \delta,I, \{(J_i,K_i)\mid i =1,\ldots, m\} \rangle$,
  the \emph{product \ac{mdp}} is $ \mathcal{M}= M \ltimes
  \mathcal{A}_\varphi =\langle V, \Sigma, \Delta, \bm D_0,\bm
  r,\gamma, \mathsf{Acc} \rangle $, with components defined as
  follows: $ V =S\times Q $ is the set of states.  $\Sigma$ is the set
  of actions.  $\bm D_0:V\rightarrow [0,1]$ is the initial
  distribution, defined by $\bm D_0((s,q))= D_0(s)$ where
  $q=\delta(I,L(s))$.  $\Delta: V\times \Sigma \times V\rightarrow
  [0,1]$ is the transition probability function. Given $v= (s,q)$,
  $\sigma$, $v'=(s',q')$ and $q'= \delta(q, L(s'))$, let $\Delta
  (v,\sigma,v')= P(s,\sigma,s')$.  The reward function is defined as
  $\bm r: V\times \Sigma \times V \rightarrow \mathbb{R}$ where given
  $v=(s,q)$, $v'=(s',q')$, $a\in \Sigma$, $\bm
  r(v,a ,v')= r(s, a, s')$ for $a\in \Sigma$.  The
  acceptance condition is $\mathsf{Acc}=\{(\hat{J_i}, \hat{K_i}) \mid
  \hat{J_i} =S\times J_i, \hat{K_i}=S\times K_i , i =1,\ldots,
  m\}$. $\blacksquare$
\end{definition}

The problem of maximizing the probability of satisfying the \ac{ltl}
formula $\varphi$ in $M$ is transformed into a problem of maximizing
the probability of reaching a particular set in the product \ac{mdp}
$\calM$, which is defined next.

\begin{definition} \cite{de1997formal} The \emph{end component} for
  the product \ac{mdp} $\mathcal{M}$ is a pair $(W, f)$ where
  $W\subseteq V $ is a non-empty set of states and $f: W \rightarrow
  \mathcal{D}(\Sigma)$ is a randomized policy. Moreover, the policy
  $f$ is defined such that for any $v\in W$, for any $a \in
  \supp(f(v))$, $\sum_{v'\in W}\Delta(v, a, v')=1$; and the induced
  directed graph $(W, \rightarrow_f)$ is strongly connected. Here,
  $v\rightarrow_f v'$ is an edge in the directed graph if $\Delta(v,
  a, v') >0$ for some $a \in \supp(f(v))$.  An \ac{aec} is an end component
  such that $W\cap \hat{J_i} =\emptyset$ and $W\cap \hat{K_i} \ne
  \emptyset$ for some $i \in \{1,\ldots, m\}$. $\blacksquare$
\end{definition}
Let the set of \ac{aec}s in $\calM$ be denoted $\AEC(\calM)$ and the
set of \emph{accepting end states} be denoted by $\calW= \{v \mid
\exists (W,f)\in \AEC(\calM), v\in W\}$. Note that, by definition, for
each \ac{aec} $(W,f)$, by exercising the associated policy $f$, the
probability of reaching any state in $W$ is 1. Due to this property,
once we enter some state $v \in \calW$, we can find at least one
accepting end component $(W,f)$ such that $v\in W$, and initiate the
policy $f$ such that for some $i\in \{1,\ldots, m\}$, all states in
$\hat{J_i}$ will be visited only a finite number of times and some
state in $\hat{K_i} $ will be visited infinitely often. The set
$\AEC(\calM)$ can be computed by algorithms
\cite{de1997formal,chatterjee2013symbolic} in polynomial time in the
size of $\calM$.

 \section{Modeling Human-in-the-loop stochastic system}

 We aim to synthesize a shared autonomy policy that switches control
 between an operator and an autonomous controller. The stochastic
 system controlled by the human operator and the autonomous
 controller, gives rise to two different \ac{mdp}s with the same set
 of states $S$, the same set $\calAP$ of atomic propositions and the
 same labeling function $L:S\rightarrow 2^\calAP$, but possibly
 different sets of actions and transition probability functions.

\begin{itemize}
\item Autonomous controller: $M_A=\langle S, \Sigma_A, T_A, \calAP, L \rangle$ where
  $T_A: S\times \Sigma_A\times S\rightarrow [0,1]$ is the transition
  probability function under autonomous controller.
\item Human operator: $M_H=\langle S, \Sigma_H, T_H, \calAP, L\rangle$  where
  $T_H: S\times \Sigma_H\times S\rightarrow [0,1]$ is the transition
  probability function under human operator.
\end{itemize}

Let $D^M_0:S\rightarrow [0,1]$ be the initial distribution of states,
same for both $M_A$ and $M_H$.  For the same system, the set of
physical actions can be the same for both the autonomous controller
and the human. We can add subscript to distinguish whose action it
is. The models $M_A$ and $M_H$ can be constructed either from prior
knowledge or from experiments by applying a policy that samples each
action from each state a sufficient amount of times
\cite{henriques2012statistical}.
% In the shared autonomous system, we define a new set of actions that
% includes the union of autonomous operator's actions and human
% operator's actions and a new state variable representing the level
% of attention paid by the human to the autonomous operator. We assume
% a finite set $\att= \{0,1,\ldots, n\}$ of different levels of
% attentions for human operator, from level $0$ to level $n$. This
% model of human's situation awareness can be generalized, see
% Remark~\ref{rmk:attlevel}. The system can switch control from the
% autonomous operator to the human operator only at level $n$ and from
% the human operator to the autonomous operator at any level.  During
% control execution, the autonomous operator can request a change in
% the level of attention and the human will grant its request
% according to a given cognitive model in the form of an \ac{mdp} as
% follows.

In the shared autonomy system, the interaction between the autonomous
controller and the operator is often made through a dialogue system
\cite{goodrich2007human}. The controller may send a request of
attention, or some other signal to the operator. The operator may
grant the request, or respond to signals, depending on his current
workload, level of attention. Admitting that it is not possible to
capture all aspects of an operator's cognitive states, we have the
following model to capture the evolution of the modeled cognitive
state.
\begin{definition}
  The operator's cognition in the shared autonomy system is modeled as
  an \ac{mdp} \[M_\att= \langle H, E, D_0^H, T_\att, \cost,\gamma, H_s
  \rangle\] where $H$ represents a finite set of cognitive states. $E$
  is a finite set of events that trigger changes in cognitive
  state. $D_0^H: H \rightarrow[0,1]$ is the initial distribution.
  $T_\att : H \times E \times H \rightarrow [0,1]$ is the transition
  probability function. $\cost: H\times E \times H\rightarrow
  \mathbb{R}$ is the cost function. $\cost(h, e, h')$ is the cost of
  human effort for the transition from $h$ to $h'$ under event
  $e$. $\gamma \in (0,1)$ is the discount factor. $H_s \subseteq H$ is
  a subset of states at which the operator can take over
  control. $\blacksquare$
\end{definition}
This cognitive model can be generalized to accomodate different model
of operator's interaction with the autonomous controller. The set $E$
of events can be requests sent by the autonomous controller to the
operator, a workload that the operator assigns to himself, or any
other external event that influences the operator's cognitive state.
This model generalizes the model of operator's cognition in
\cite{Mouaddib2010}, in which an event is a request to increase,
decrease, or maintain the operator's attention in the control task. In
particular, it is assumed that in a particular set of states,
transitions from the autonomous controller to the operator can
happen. For instance, for tele-operated robotic arm or semi-autonomous
vehicle, operator may take over control only when he is aware of the
system's state and not occupied by other tasks
\cite{goodrich2007human}.  The model is flexible and can be extended to other cognitive models in
shared autonomy.  In this
paper, we assume the model of operator for the given task is
given. One can obtain such a model by statistical learning
\cite{rothkopf2013modular}.

We illustrate the concepts using the robotic arm example.
\begin{example}
  Consider a robot manipulator having to pick up the objects on a
  table and place it into a box. There are two types of objects, small
  and large. For small and large objects, the probabilities of a
  successful pick-and-place maneuver performed by the autonomous
  controller is $85\%$ and $50\%$ respectively. The \ac{mdp} for the
  controller is shown in Figure~1a. With an operator
  tele-operating the robot, the probabilities of a successful
  pick-and-place maneuver is $95\%$ and $75\%$ respectively. The
  operator's cognitive model includes two cognitive states: $0$
  represents the state when the human does not pay any attention to
  the system (at the attention level $0$), and $1$ represents the
  state when he pays full attention (at the attention level $1$). The
  set of events in $M_\att$ is the requests of human attention to the
  task, $E=\{0,1\}$ where $e \in E$ represents the current requested
  attention level is $e$. For any $h \in \{0,1\}$, $e \in E$, let
  $\cost(h,e, h')=10$ for $h'=1$, otherwise $5$. The transition
  probability function of $M_\att$ is shown in
  Figure~1b.

\begin{figure}[ht]
\centering
\includegraphics[width=0.45\textwidth]{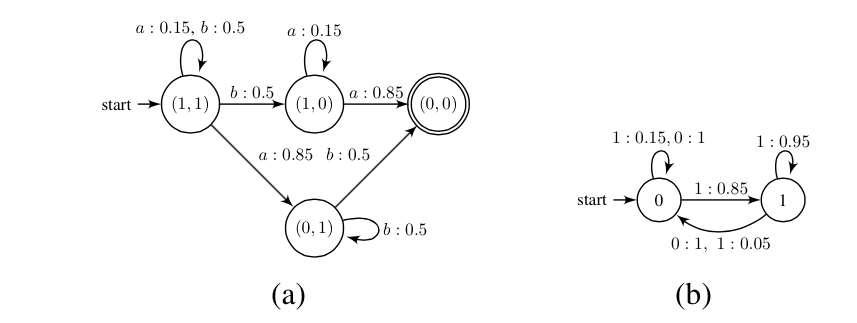}
\caption{(a) The \ac{mdp} for the robotic arm controlled by the
  autonomous controller. A state $(n,m)$ represents there are $n$ small
  objects and $m$ large objects remaining to be picked.  The available
  actions are $a$ and $b$ for picking up small and large objects,
  respectively. The \ac{mdp} for the robotic arm tele-operated by the
  human can be obtained by changing the probabilities on the
  transitions. (b) The \ac{mdp} $M_\att$ for modeling the dynamics of
  human's attention changes.\label{}}
\end{figure}
\end{example}

Given two \ac{mdp}s, $M_A$ for the controller and $M_H$ for the
operator, and a cognitive model for the operator $M_\att$, we
construct a \emph{shared autonomy stochastic systems} as an \ac{mdp}
as follows.
% \ac{mdp} that captures the dynamics of the system under
% the switching control of human and autonomous controller, as well as
% the cognitive model of the operator as follows.
\[
M_{SA}= \langle \bm S , \Sigma, \bm T,   D_0 , \calAP, L, \cost, \gamma \rangle
\]
where $ \bm S= S\times H$ is the set of states. A state $(s,h)$
includes a state $ s$ of the system and a cognitive state $h$ of
human.  $\Sigma = (\Sigma_A \cup \Sigma_H) \times E $ is the set of
actions. If $(a, e ) \in \Sigma_A\times E$, the system is controlled
by the autonomous controller and the event affecting human's cognition
is $e$. If $(a,e) \in \Sigma_H \times E$, the system is controlled by
human operator and the event affecting human's cognition is $e$.  $\bm
T :\bm S\times \Sigma \times \bm S\rightarrow [0,1] $ is the
transition probability function, defined as follows. Given a state
$(s, h)$ and action $(a,e)\in \Sigma_A\times E$, $\bm T((s,h),(a,e),
(s',h'))= T_A(s,a,s') T_\att(h, e, h')$, which expresses that the
controller acts and triggers an event that affects the operator's
cognitive state.  Given a state $(s,h)$ for $h \in H_s$, and action
$(a,e)\in \Sigma_H\times E$, $\bm T ((s,h), (a, e ), (s', h'))= T_H(q,
a, q') T_\att(h,e, h')$, which expresses that the operator controls
the system and an event $e$ happens and may affect the cognitive
state.  $ D_0: S\times H \rightarrow[0,1] $ is the initial
distribution. $ D_0(s,h)= D^M_0(s) \times D^H_0(h)$, for all $s\in S$,
$h\in H$.  $L: \bm S\rightarrow 2^\calAP$ is the labeling function
such that $L((s,h))=L(s)$. $ \cost: \bm S \times \Sigma \times \bm
S\rightarrow R $ is a cost function for human effort defined over the
state and action spaces and $ \cost ((s,h), (a, e),(s',h'))=
\cost(h,e, h')$. $\gamma \in (0,1)$ is the discount factor, the same
in $M_\att$.  Slightly abusing the notation, we denote the cost
function in $M_{SA}$ the same as the cost function in $M_\att$ and the
labeling function in $M_{SA}$ the same as the labeling function in
$M$.

Note that, although the cost of human effort only contains the cost in
his cognitive model, it is straightforward to incorporate the cost of
human's actions into the cost function.

 \addtocounter{example}{-1}
 \begin{example}(Cont.) We construct \ac{mdp} $M_{SA}$ in
   Figure~\ref{fig:samdp} for the robotic arm example. For example, $\bm
   T((1,1),0), (a_A,1), ((0,1),1))= T((1,1), a_A, (0,1)) \cdot
   T_\att(0,1,1)=0.75\cdot 0.85=0.7225$, which means the probability
   of the robot successfully picking up a small object and placing it
   into the box while the human changes his cognitive state to 1
   (fully focused) upon the robot's request is $0.7225$. Also it is
   noted that from the states $((0,1),0)$ and $((1,1),0)$, no human's
   action is enabled.  The cost function is defined such that
   $\cost((q,h),a, (q,h'))=10$ if $h'=1$, otherwise $5$.
\vspace{-2ex}
    \begin{figure}[ht]
\centering
 \includegraphics[width=0.45\textwidth]{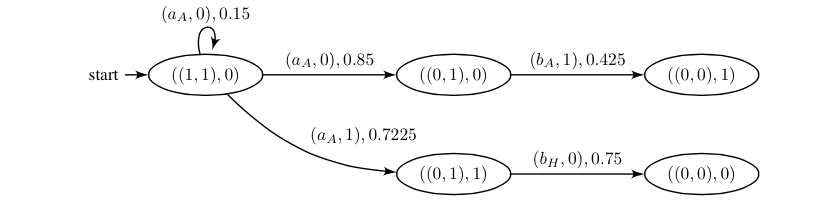}
 \caption{A fragment of \ac{mdp} $M_{SA}$ for robotic arm example
   (note only a subset of states and transitions are shown).
   Subscripts $A$ and $H$ distinguish actions performed by the
   autonomous controller ($A$) and the human ($H$), respectively. }
\label{fig:samdp}
\end{figure}
\vspace{-2ex}
\end{example}
The main problem we solve is the following.
\begin{problem}
\label{def:problem}
  Given a stochastic system under shared autonomy control between an
  operator and an autonomous controller, modeled as \ac{mdp}s $M_H$ and
  $M_A$, a model of human's cognition $M_\att$, and an \ac{ltl}
  specification $\varphi$, compute a policy that is Pareto optimal
  with respect to two objectives: \begin{inparaenum}[1)]\item Maximizing the
    discounted probability of satisfying the \ac{ltl} specification
    $\varphi$ and \item minimizing the discounted total cost of human
    effort over an infinite horizon. \end{inparaenum}  
\end{problem}
The definition of Pareto optimality in this context is given formally
at the beginning of section~\ref{sec:multiobj}. By following a Pareto
optimal policy, we achieve a balance between two objectives: It is
impossible to make one better off without making the other one worse
off.

\section{Synthesis for shared autonomy policy}
Given an \ac{mdp} $M_{SA} =\langle \bm S, \Sigma, \bm T, D_0,
\calAP,L, \cost,\gamma \rangle$ and a \ac{dra} $\mathcal{A}_\varphi=
\langle Q ,2^{\calAP}, \delta, I, \{(J_i,K_i)\mid i =1,\ldots, m\}
\rangle$, the product \ac{mdp} following Definition~\ref{def:product}
is $ \mathcal{M}= M_{SA} \ltimes \mathcal{A}_\varphi =\langle V,
\Sigma, \Delta, \bm D_0, \cost^{\calM},\gamma, \mathsf{Acc} \rangle
$. Recall that the policy maximizing the probability of satisfying the
\ac{ltl} specification is obtained by first computing the set of
\ac{aec}s in $\calM$ and then finding a policy that maximizing the
probability of hitting the set $\calW$ of states contained in
\ac{aec}s (see Section~\ref{sub:mdpltl}).

For quantitative \ac{ltl} objectives, for example, maximizing the
probability of satisfying an \ac{ltl} formula, or a discounted reward
objective over an infinite horizon, a memoryless policy in the product
\ac{mdp} suffices for optimality
\cite{baier2008principles,Filar1996}. % Note that a memoryless policy in
% the product \ac{mdp} can be transformed into a finite-memory policy
% in the \ac{mdp} where the set $Q$ in the specification \ac{dra} is
% the set of memory states.
In the following, by policies, we mean
memoryless ones in the product
\ac{mdp}. 

% In the case of \ac{mdp}s, not only we need to optimize a quantitative
% criterion with respect to the given temporal logic objective, but also
% we want to minimize the cost of human effort. Generally, the policy
% which is optimal with respect to the former might not be optimal with
% respect to the latter. Thus,
Problem~\ref{def:problem} is in fact a multi-objective optimization
problem for which we need to balance the cost of human's effort and
satisfaction for \ac{ltl} constraints. However, the solutions for
multi-objective \ac{mdp}s cannot be directly applied due to the
constraint that once the system runs into an \ac{aec} of $\calM$, the
policy should be constrained such that all states in that \ac{aec} are
visited infinitely often. Based on the particular constraint, we
divide the original problem into a two-stage optimization problem: The
policy synthesis for \ac{aec}s is separated from solving a
multi-objective \ac{mdp} formulated before reaching a state in an
\ac{aec}.

\subsection{Pareto efficiency before reaching the \ac{aec}s}
\label{sec:multiobj}
The first stage is to balance between a quantitative criterion for a
temporal logic objective and a criterion with respect to the cost of
human effort before a state in the set $\calW$ is reached. Remind that
$\calW$ is the union of states in the accepting end components of
$\calM$.  We formulate it as an \ac{momdp}.  However, for objectives
of different types, such as, discounted, undiscounted, and
limit-average. the scalarization method for solving \ac{momdp}s does
not apply. Thus, we consider to use the discounted reachability
property \cite{de2005model} for the given \ac{ltl} specification, as
well as discounted costs for the human attention, with the same
discount factor $\gamma\in (0,1) $ specifying the relative
importance of immediate rewards.

For an \ac{ltl} specification, discounting in the state sequence
before reaching the set $\calW$ means that the number of steps for
reaching $\calW$ is concerned \cite{de2005model}. Without discounting, as
long as two policies have the same probability of reaching the set
$\calW$, they are equivalent regardless of their expected numbers of
steps to reach $\calW$. With discounting though, a policy has smaller expected
number of steps in reaching $\calW$ is considered to be better than
the other. 

% For convenience, we transform cost into reward by letting
% $r(v,a,v')=-\cost(v,a,v')$. Thus, instead of minimizing the cost, we
% maximize the expectation of discounted total reward for human
% attention. The following definition is adopted from
% \cite{chatterjee2006markov}.
\begin{definition}
\label{def:momdpvaluefunc}
Given the product \ac{mdp} $\calM$, for a state $v$ in $\calM$, the
\emph{discounted probability} for reaching the set $\calW$ under
policy $f: V\setminus \calW\rightarrow \calD(\Sigma)$ is
\[U_1 (v,f)= 
 \bbE^f_v\left[\sum_{t=0}^\infty \gamma^t \cdot  r_1(X_t,A_t, X_{t+1})\right]
\]
where the reward function $r_1: V\times \Sigma \times V
\rightarrow \{0,1\}$ is defined such that $r_1(v,a,
v')=1$ if and only if $v\notin \calW$ and $v'\in \calW$, otherwise
$r_1(v,a, v')=0$.  The discounted total reward with respect
to human attention for a policy $f: V\setminus \calW\rightarrow
\calD(\Sigma)$ and a state $v$ is
\[ U_2(v,f)= \bbE^f_v\left[ \sum_{t=0}^\infty \gamma^t \cdot r_2(X_t,
  A_t, X_{t+1}) \right],\] where the reward function $r_2: V\times
\Sigma \times V\rightarrow \mathbb{R}$ is defined such that $r_2(v,a,
v')= - \cost^{\calM}(v,a, v')$ if and only if $v \notin \calW$ and $v' \notin
\calW$, $r_2(v,a, v')= - U_{\AEC}^\ast(v')$ if $v\notin \calW$ and
$v'\in \calW$, and $r_2(v,a,v')=0$ otherwise. Here,
$U_{\AEC}^\ast:\calW\rightarrow \mathbb{R}$ is the discounted cost of
human attention for remaining in an accepting end components under the
optimal policy for the second stage. $\blacksquare$
\end{definition}
The \emph{discounted value profile}, at $v$ for policy $f$, is defined
as $\vec{U}(v, f)= ( U_1(v,f),U_2 (v,f))$. We denote $\vec{r}=
(r_1,r_2)$ as the vector of reward functions.  The function
$U_{\AEC}^\ast:\calW\rightarrow \mathbb{R}$ is computed in the next
section.

\begin{definition}\cite{chatterjee2006markov}
  Given an \ac{mdp} $\calM =\langle V, \Sigma, \Delta \rangle$ and a
  vector of reward functions $\vec{r}= (r_1,r_2,\ldots, r_n)$, for a
  given state $v\in V$, policy $f$ \emph{Pareto-dominates} policy $f'$
  at state $v$ if and only if $ \vec{U}(v,f ) = (U_1(v,f),\ldots,
  U_n(v,f))\ne \vec{U}(v,f') = (U_1(v,f'),\ldots, U_n(v,f')) $ and for
  all $ i=1,\ldots, n, U_i(v,f) \ge U_i(v,f') $.  A policy $f$ is
  \emph{Pareto optimal} in a state $v \in V$ if there is no other
  policy $f'$ Pareto-dominating $f$. For a Pareto-optimal policy $f$
  at state $v$, the corresponding value profile $\vec{U}(v,f)$ is
  referred to as a \emph{Pareto-optimal point (or an efficient
    point)}. The set of Pareto-optimal point are called the
  \emph{Pareto set}. $\blacksquare$
\end{definition} A Pareto optimal policy $f$ for a given initial
distribution is defined analogously by comparing the expectations of
value functions under the initial distribution. 

We employ Tchebycheff scalarization method
\cite{steuer92b,perny2010finding} to find Pareto optimal policies for
user specified weights. First, we solve a set of single objective
\ac{mdp}s, one for each reward function. Let $U_i(\cdot, f_i^\ast):
V\rightarrow \mathbb{R}$ be the value function of the optimal policy
$f_i^\ast$ with respect to the $i$-th reward function. The ideal point
$U^I=(U_1^I, U_2^I)$ is then computed as follows: for $i=1,2$,
$U_i^I=\sum_{v\in V} \bm D_0(v)U_{i}(v,f_i^\ast)$. Given a weight
vector $\vec{w} = (w_1, w_2)$ where $w_i$ is the weight for the $i$-th
criterion such that $w_1+w_2=1$, a Pareto optimal policy associated
with the weight vector $\vec{w}$ can be found with the following
nonlinear program:

\begin{align}
\small
\begin{split}
  & \min_x \max_{i=1,2} (\lambda_i \cdot ( U^I_i -R_i\cdot x))+
  \epsilon\sum_{i=1,2} \lambda_i \cdot \left(U^I_i -
    R_i\cdot x\right) \\
  & \text{subject to: }  \forall  v\in V \setminus\calW,\\
  &\sum_{a\in \act(v)} x(v,a)=\bm D_0(v) + \gamma \sum_{v'\in V}
  \sum_{a'\in \act(v')} \Delta(v',a', v) \cdot x(v', a'), \\
  & \text{and }\forall v\in V\setminus \calW, \forall a \in \Sigma,
  \quad x(v,a) \ge 0, \quad
\end{split}
\label{eq:linearprog}
\end{align}
where $\epsilon$ is a small positive real that can be chosen
arbitrarily, $x(v,a)$ is interpreted as the expected discounted
frequency of reaching the state $v$ and then choosing action $a$,
$R_i\cdot x= \sum_{v\in V}\sum_{a\in \act(v)} \sum_{v'\in V}
r_i(v,a,v') \Delta(v,a,v') x(v,a)$, and $\vec{\lambda}$ is a positive
weighting vector computed from a weight $\vec{w}$, the ideal points
and the Nadir points \cite{steuer92b} for all reward functions
(detailed in Appendix).  The nonlinear programming problem can then be
formulated into a linear programming problem in the standard way by
setting a new variable $z =\max_{i=1,2} (\lambda_i \cdot ( U^I_i -R_i\cdot
x)) $.  % as follows.
% \begin{equation}
% \small
% \begin{split}
% &  \min_{z,x} z+  \epsilon \sum_{i=1,\ldots, n} \lambda_i( U^I[i]  -  R_i\cdot x)\\
% & \text{subject to: }  \forall v\in V\setminus \calW, \\
%  &\sum_{a\in \act(v)} x(v,a)= \bm D_0(v) +\gamma \sum_{v'\in V}
%   \sum_{a'\in \act(v')} \Delta(v',a', v) \cdot x(v', a'), \\
%   &\forall i=1,\ldots, n , z\ge \lambda_i (U^I[i]-R_i\cdot x), \\
%  & \text{ and }\forall v\in V\setminus \calW, \forall a \in
%   \Sigma, x(v,a) \ge 0. 
% \end{split}
% \label{eq:linearprog}
% \end{equation}
%  Let $\{x(v,a)\mid v\in V,
% a\in \act(v)\}$ be the solution to \eqref{eq:linearprog}.  
The Pareto
optimal policy $f:V\rightarrow \calD(\Sigma)$ is defined such that
\begin{equation}
\label{eq:obtainPareto}
f(v)(a)= \frac{x(v,a)}{\sum_{a\in \act(v)}x(v)},
\end{equation}
which selects action $a$ with probability $f(v)(a)$ from the state $v$,
for all $v\in V$, $a\in \act(v)$.

\begin{example}
  Continue with the robot arm example. Given the discount factor
  $\gamma=0.98$, for the simple objective ($1$st objective) as quickly
  as possible of reaching a state at which all objects are in the box,
  the optimal strategy $f_1^{\ast}$ is shown in the first row of
  Table~\ref{tbl:pickplacePareto}. Intuitively, the robot starts by
  requesting the operator to increase his level of attention
  and % at the initial state,
  % the robot attempts to pick up the small object and sends the
  % operator a request to increase his level of attention to $1$. If
  % the
  % request is granted, it will switch to the control to the operator
  % to
  % pick up all the remaining objects. Otherwise, it will continue to
  % request the level of attention to be $1$ while attempting to pick
  % up
  % the small object if the previous attempt fails or the large object
  % if the small one is already in the box. In other words,
  wants to switch control to human as soon as possible as the latter
  has higher probability of success for a pick-and-place
  maneuver. Alternatively, the optimal policy with respect to
  minimizing the cost of human effort ($2$nd objective), is to let the
  robot pick up all the objects since by doing so, eventually all the
  objects will be collected into the box. The strategy $f_2^\ast$ is
  shown in the second row of Table~\ref{tbl:pickplacePareto}.

  Now suppose that a user gives a weight $0.8$ for the first objective
  and $0.2$ for the second objective, through normalization, the new
  weight vector $\vec{\lambda}= (11.93,0.02)$, is obtained with the
  method in Appendix. By solving the linear programming problem in
  \eqref{eq:linearprog}, we obtain a Pareto-optimal policy $f_P^\ast$
  shown in the third row of Table~\ref{tbl:pickplacePareto}. Noting
  that the difference of $f_P^\ast$ and $f_1^\ast$ is that when it
  comes to the small object, if the current human attention is high,
  the robot will request the human to decrease his attention level and
  therefore, if the object fails to be picked up through
  tele-operation, the autonomous controller will take over for picking
  up the small object. Whileas in $f_1^\ast$, the robot prefers the
  human operator to pick up all objects, no matter it is a big one or
  a small one.

Figure~\ref{fig:pickplacePareto}
  shows the state value for the initial state $v_0 = ((1,1),0)$ with
  respect to reward functions $r_1,r_2$, under the policies
  $f_1^\ast$, $f_2^\ast$ and a subset of Pareto optimal policies, one
  for each weight vector $\vec{w}$ in the set $\{(\beta,1-\beta)\mid
  \beta =\frac{k}{10}, k =1,2,\ldots,9\}$.

\begin{table}[ht]\centering
\caption{Policies for pick-and-place task}
\scalebox{0.6}{
\begin{tabular}{l | cccccc}
States:& $((1, 1), 0) $ & $((1, 1), 1) $ & $((1, 0), 0)$ & $((1,0),1)$ & $((0, 1),
1) $ & $((0, 1), 0) $ \\
\hline
 $f_1^\ast$: & $(a_A,1)$ & $(b_H,1)$ & $(a_A,1)$ & $(a_H,1)$& $(b_H,1)$ & $(b_A,1)$ \\
$f_2^\ast$ : &  $(a_A,0)$ & NA & $(a_A,0)$ & NA & NA & $(b_A,0)$\\
$f_P^\ast$ & $(a_A,1)$ & $(b_H,1)$ & $(a_A,0)$ & $(a_H,0)$ & $(b_H,1)$ & $(b_A,1)$
\end{tabular}}
\label{tbl:pickplacePareto}
\end{table}
\vspace{-6ex}

\begin{figure}[ht]
 \centering
\includegraphics[width=0.5\textwidth]{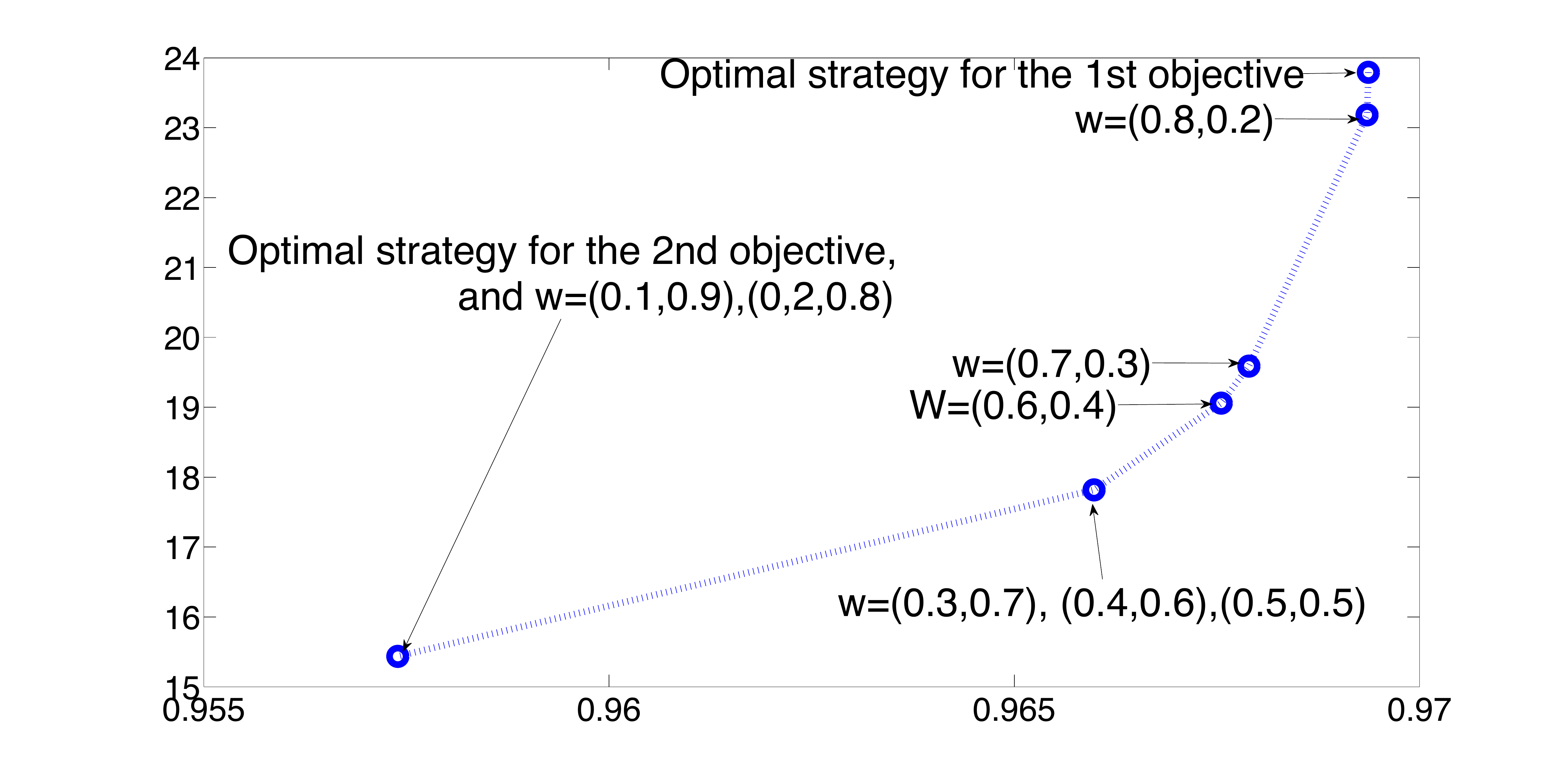}
\caption{The state values of the initial state with respect to reward
  functions $r_1,r_2$, under policies $f_1^\ast$, $f_2^\ast$ and a set
  of Pareto optimal policies $f_P^\ast$, one for each weight vectors
  in the set $\{(\beta,1-\beta)\mid \beta =\frac{k}{10}, k
  =1,2,\ldots,9\}$. The $x$-axis and $y$-axis represent the values of
  the initial state under the 1st and 2nd criteria, respectively.}
\label{fig:pickplacePareto}
\end{figure}
Though the Pareto optimal policy for $\vec{w}=(0.8,0.2)$ is
deterministic in this example. It may generally need to be randomized
for a given weight vector.
\end{example}

So far we have introduced a method for synthesizing Pareto optimal
policies before reaching a state in one of the accepting end components. Next, we
introduce a constrained optimization for synthesizing a policy that
minimize the expected discounted cost of staying in an \ac{aec} and visiting
all the states in that \ac{aec} infinitely often.

\subsection{A constrained optimization for accepting end components}

For a state $v$ in $\calW$, one can identify at least one \ac{aec}
$(W,f)$ such that $v\in W$. It is noted that the policy $f:
V\rightarrow \mathcal{D}(\Sigma)$ is a randomized policy that ensures
every state in $W$ is visited infinitely often with probability 1
\cite{chatterjee2013symbolic}.  However, there might be more than one
\ac{aec} that contains a state $v$, and we need to decide which
\ac{aec} to stay in such that the expected discounted cost of human
effort for the control execution over an infinite horizon is
minimized. % Informally, the minimal discounted total cost of staying in
% an \ac{aec} after visiting a state $v\in \calW$ becomes the terminal
% cost of the multi-objective optimization problem in the first stage.

We consider a constrained optimization problem: For each \ac{aec}
$(W,f)$ where $W\subseteq V$ and $f: W\rightarrow
\mathcal{D}(\Sigma)$, solve for a policy $g: W\rightarrow
\mathcal{D}(\Sigma)$ such that the cost of human effort for staying in
that \ac{aec} is minimized. The constrained optimization problem is
formulated as follows.
\begin{align}\small
\label{eq:optAEC}
\begin{split}
  &\min_{g} U_{\AEC}(v,g,W)=\sum_{k=0}^\infty\gamma^k \cdot
  \bbE_{v}^{g}[\cost^{\calM}(X_t,A_t, X_{t+1})]\\
&  \text{subject to: } \forall v\in W, \Pr^g(\forall t, \exists t'>t
,X_{t'}=v) =1, \text{ and }\\
  & \forall v \in W, \forall a \notin \act(v), g(v)(a)=0,
\end{split}
\end{align}
where the term $\Pr^g(\forall t, \exists t'>t ,X_{t'}=v)$ measures the
probability of infinitely revisiting state $v$ under policy $g$.

The linear program formulated for solving \eqref{eq:optAEC} can be
obtained as follows:
\begin{align}
\small
\label{eq:dualoptAEC}
\begin{split}
&\min \sum_{v\in V}\sum_{a\in \act(v)} \left[x(v,a) \cdot \left(\sum_{v'\in
    V}\cost^{\calM}(v,a,v') \Delta(v,a,v') \right)
  \right]\\
& \text{subject to:  for }  v\in W, \\ 
& \sum_{a\in \act(v)} x(v,a)= \eta(v) + \gamma
\sum_{v'\in V} \sum_{a'\in \act(v')} \Delta(v',a', v) \cdot x(v',
a'),\\
&\forall v\in W, \forall a \in
\Sigma, x(v,a) \ge 0,\\
& \forall v \in
W, \sum_{a\in \act(v)} x(v,a) >=\varepsilon, \text{ and }\\
&  \forall v \in W, \forall a \notin \act(v), x(v,a) =0,
\end{split}
\end{align}
where $\varepsilon$ is an arbitrarily small positive
real. $\eta:W\rightarrow [0,1]$ is the initial distribution of states
when entering the set $W$. Because for single objective optimization
the optimal state value does not depend on the initial distribution
\cite{puterman2009markov}, $\eta$ can be chosen arbitrarily from the
set of distributions over $W$. The physical meaning of $\sum_{a\in
  \act(v)}x(v,a)$ is the discounted frequency of visiting the state
$v$, which is strictly smaller than the frequency of visiting the
state $v$ as long as $\gamma \ne 1$. By enforcing the constraints
$\sum_{a\in \act(v)} x(v,a) >=\varepsilon $, we ensure that the
frequency of visiting every state in $W$ is non-zero, i.e., all states
in $W$ will be visited infinitely often.

The solution to \eqref{eq:dualoptAEC} produces a memoryless policy
$g^\ast : W \rightarrow \calD(\Sigma)$ that chooses action $a$ at a
state $v$ with probability $g^\ast(v)(a)= \frac{x(v,a)}{\sum_{a\in
    \act(v)}x(v,a)}$. Using policy evaluation
\cite{barto1998reinforcement}, the state value $U_\AEC^\ast(v,W)$ for
each $v\in W$ under the optimal policy $g^\ast$ can be computed.
Then, the terminal cost $U_{\AEC}^\ast: \calW\rightarrow
\mathbb{R}$ is defined as follows.
\[
U_{\AEC}^\ast (v) = \min_{(W,f)\in \AEC} U_{\AEC}^\ast(v,W)
\]
and the policy after hitting the state $v$ is $g$ such that $U_{\AEC}^g(v,W)=U_{\AEC}^\ast(v,W) =U_{\AEC}^\ast(v)$.

We now present Algorithm~\ref{alg:twostage} to conclude the two-state
optimization procedure.

% \begin{theorem}

% \end{theorem}
\begin{algorithm}[h]
  \SetKwFunction{GetSAMDP}{GetProductMDP}
  \SetKwFunction{GetAEC}{GetAEC}
  \SetKwFunction{ConstrainedOptAEC}{ConstrainedOptAEC}
  \SetKwFunction{PolicyEvaluate}{PolicyEvaluate}
  \SetKwFunction{GetRewardVec}{GetRewardVec}
  \SetKwFunction{GetParetoOptimal}{GetParetoOptimal} 
% \SetKwComment{Comment}{\/\*}{\*\/}
  { \KwIn{The \ac{mdp} $M_A, M_H$ and $M_\att$, a specification
      automaton \ac{dra} $\mathcal{A}_\varphi$, and a weight
      $\vec{w}$.} \KwOut{A pareto policy $f$ for the discounted
      reachability and a partial function $\mathsf{Policy}:
      V\rightarrow \calF$, where $\calF$ is the set of randomized
      policies. $\mathsf{Policy}(v)$ is the policy to follow after
      state $v$ is reached.}
\Begin
{
$\calM=$ \GetSAMDP$(M_A,M_H,M_\att, \calA_\varphi)$\;
$\AEC(\calM)=$ \GetAEC$(\calM)$ \tcc*{Compute the accepting end
  components.}
\For{$(W,f)\in \AEC$}{$g^\ast_W=$\ConstrainedOptAEC$(W, \cost^{\calM})$\;
  \tcc{Solve \eqref{eq:dualoptAEC}.}
$U_\AEC^\ast(v,W)=$\PolicyEvaluate$(W,\cost^{\calM},g^\ast_W)$\;
}
$\calW=\cup_{(W,f)\in\AEC} W$\;
\For{$v\in \calW$}{$U_\AEC^\ast(v)=\min_{(W,f)\in \AEC}
  U_\AEC^\ast(v,W)$\;
$\mathsf{Policy}(v)= g^\ast_W$ for which $W$ such that $ U_\AEC^\ast(v,W) =
U_{\AEC}^\ast(v)$.}
$\vec{r}=$ \GetRewardVec$(\calM,\{ U_\AEC^\ast(v) \mid v\in \calW\},
\calW )$\; \tcc{Formulate the reward vector according to
  Definition~\ref{def:momdpvaluefunc}.}
$f=$\GetParetoOptimal$(\vec{r},\calM,\vec{w})$
\tcc{Solve \eqref{eq:linearprog} and obtain the Pareto optimal policy $f$ as in \eqref{eq:obtainPareto}.}
\Return $f, \mathsf{Policy}$.}
}
\caption{\texttt{TwoStageOptimization}
\label{alg:twostage}}
\end{algorithm}

\paragraph*{Remark}
{% Note that because of the separation of the probability for
 %  satisfying the \ac{ltl} specification and the cost of human effort,
 %  it is thus possible to modify the cost function in order to provide
 %  incentive for other desirable behavior to emerge while the
 %  satisfaction of the \ac{ltl} specification in the \ac{aec}s is
 %  unaffected. For example, given a specification $\square \lozenge
 %  R_1$, which requests the system to always eventually visit a state
 %  at which $R_1$ evaluated true, if we add additional reward to states
 %  at which $R_1$ is true, then the policy derived from the second
 %  stage will have a higher frequency of visiting such a state than a
 %  policy synthesized without this additional reward. Moreover,
  Although in this paper we only considered two objectives, the
  methods can be easily extended to more than two objectives for
  handling \ac{ltl} specifications and different reward/cost
  structures in synthesis for stochastic systems, for example, the
  objective of balancing between the probability of
  satisfying an \ac{ltl} formula, the discounted total cost of human
  effort, and the discounted total cost of energy consumption.}

\section{An example on shared autonomy}
We apply Algorithm~\ref{alg:twostage} to a robotic motion planning
problem in a stochastic environment. The implementations are in
Python and Matlab on a desktop with Intel(R) Core(TM) processor and 16
GB of memory.

Figure~4a shows a gridworld environment of four
different terrains: Pavement, grass, gravel and sand. In each terrain,
the mobile robot can move in four directions (heading north `N', south
`S', east `E', and west `W'). There is onboard feedback controller
that implements these four maneuver, which are motion primitives.
Using the onboard controller, the probability of arriving at the
correct cell is $95\%$ for pavement, $80\%$ for grass, $75\%$ for
gravel and $65\%$ for sand. Alternatively, if the robot is operated a
human, it can implement the four actions with a better performance for
terrains grass, sand and gravel.  The probability of arriving at the
correct cell under human's operation is $95\%$ for pavement, $90\%$
for grass, $85\%$ for gravel and $80\%$ for sand. The objective is
that either the robot has to visit region $R_1$ and then $R_2$, in
this order, or it needs to visit region $R_3$ infinitely often, while
avoiding all the obstacles.  Formally, the specification is expressed
with an \ac{ltl} formula $\varphi = \left( \lozenge (R_1 \land
  \lozenge R_2 ) \lor \square \lozenge R_3 \right) \land \square
\lozenge \neg \text{Unsafe}$.

Figure~4b is the cognitive model of the
operator, including three states : $L$, $M$ and $H$ represent that
human pays low, moderate, and high attention to the system
respectively. The costs of paying low, moderate and high attention to
the system are $1$, $5$, and $10$, respectively. Action `$+$'
(resp. '$-$') means a request to increase (resp. decrease) the
attention and action $\lambda$ means a request to maintain the current
attention. The operator takes over control at state $H$.

\begin{figure}[ht]
\centering
\includegraphics[width=0.45\textwidth]{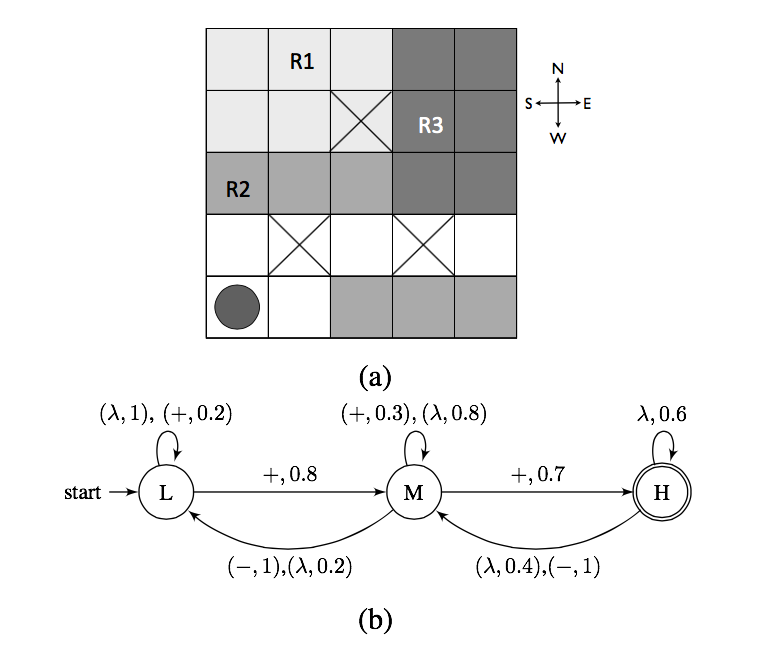}
\caption{(a) A $5\times 5$ gridworld, where the disk represents the
  robot, the cells $R_1$, $R_2$, and $R_3$ are the interested regions,
  the crossed cells are obstacles. We assume that if the robot hits
  the wall (edges), it will be bounced back to the previous
  cell. Different grey scales represents different terrains: From the
  darkest to the lightest, these are “sand,” “grass,” “pavement” and
  “gravel.” (b) The \ac{mdp} $M_\att$ of the human operator.}
\label{fig:gridworldex}
\vspace{-5ex}
\end{figure}

\begin{figure}[ht]
\centering
\includegraphics[width=0.4\textwidth]{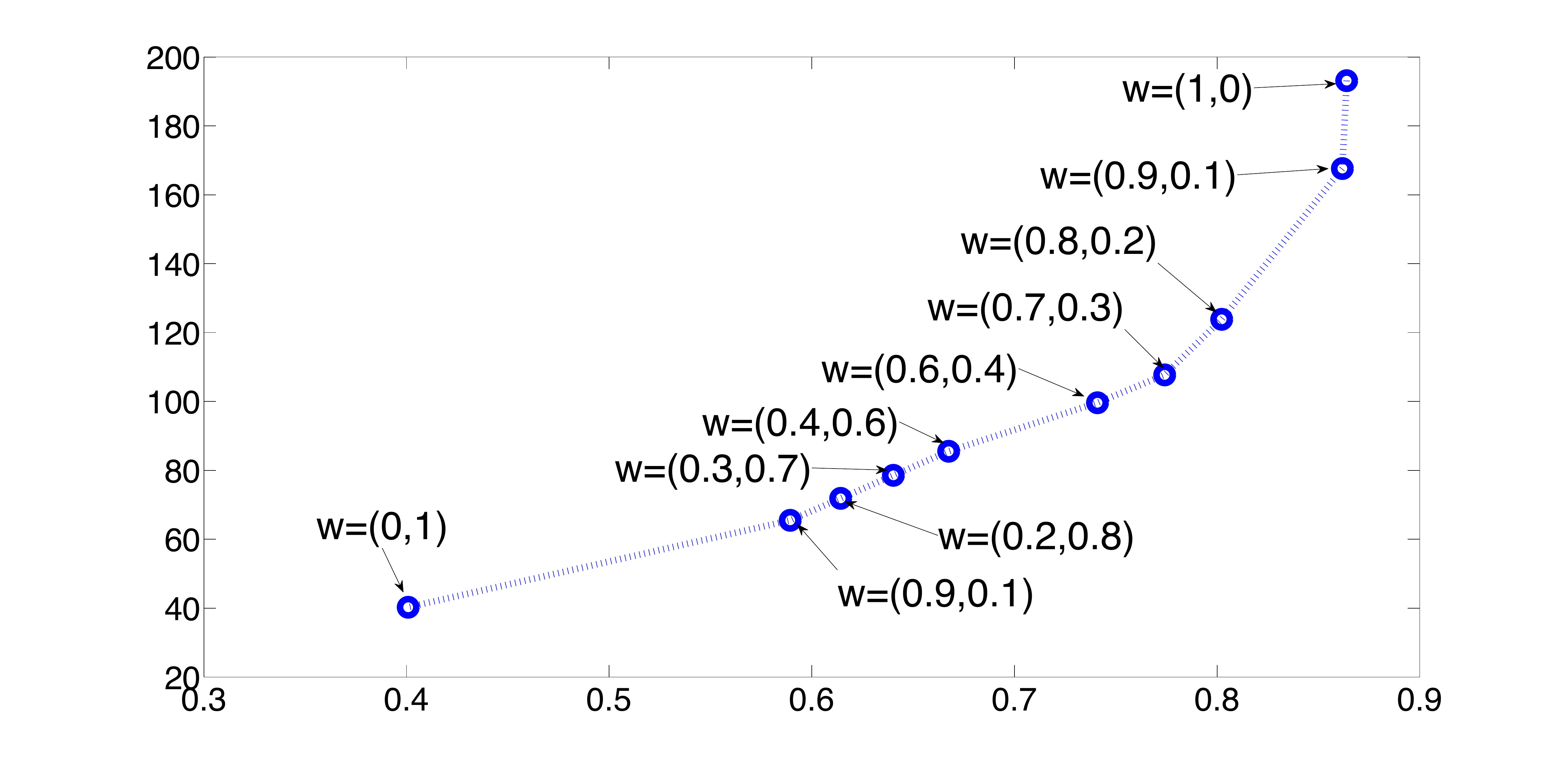}
\caption{The state values of the initial state given reward functions
  $r_1,r_2$, under policies $f_1^\ast$, $f_2^\ast$ and a set of Pareto
  optimal policies $f_P^\ast$, for each $\vec{w} \in \{(\beta,1-\beta)\mid \beta =\frac{k}{10}, k =1,2,\ldots,9\}$. The
  $x$-axis represents the values of the initial state for discounted
  probability of satisfying the \ac{ltl} specification. The $y$-axis
  represents the values of the initial state with respect to the cost
  of human effort.}
\label{fig:gridworldPareto}
\end{figure}

During control execution, we aim to design a policy that coordinates
the switching of control between the operator and the autonomous
controller, i.e., onboard software controller. The policy should be
Pareto optimal in order to balance between maximizing the expected
discounted probability of satisfying the \ac{ltl} formula $\varphi$,
and minimizing the expected discounted total cost of human
efforts. Figure~\ref{fig:gridworldPareto} shows the state value for
the initial state with respect to reward functions $r_1$ for the
\ac{ltl} formula and $r_2$ for the cost of human effort, under the
single objective optimal policy $f_1^\ast$ and $f_2^\ast$, and a
subset of Pareto optimal policies, one for each weight vectors
$\vec{w}$ in the set $\{(\beta,1-\beta)\mid \beta=\frac{k}{10},
k=1,2,\ldots, 9\}$. For the \ac{ltl} specification, all policies are
randomized.% It is noted that if the structures of \ac{mdp}s $M_A$ and $M_H$ are
% the same, then the optimal policy for the second stage, which allows
% the system to staying in an \ac{aec} and visit all
% the states infinitely often, will keep the cost of human effort to the
% minimum. If it is possible to remain in a state in $M_C$ while the
% cost of human effort is minimal for all states in $M_C$, then once the
% system enters an \ac{aec}, it will maintain he human performance
% state at this lowest cost one. Then we only need to estimate the cost
% using 

\vspace{-2ex}

\section{Concluding remarks and critiques}
We developed a synthesis method for a class of shared autonomy systems
featured by switching control between a human operator and an autonomous
controller.  In the presence of inherent uncertainties in the
systems' dynamics and the evolution of humans' cognitive states, we
proposed a two-stage optimization method to trade-off the human effort
for the system's performance in satisfying a given temporal logic
specification. Moreover, the solution method can also be extended for
solving multi-objective \ac{mdp}s with temporal logic constraints.  In
the following, we discuss some of the limitations in both modeling and solution
approach in this paper and possible directions for future work.

We employed two \ac{mdp}s for modeling the system operated by the human
and for representing the evolution of
cognitive states triggered by external events such as workload,
fatigue and requests for attention. We assumed that
these models are given. However, in practice, we might need to learn
such models through experiments and then design adaptive shared
autonomy policies based on the knowledge accumulated over the learning
phase. In this respect, a possible solution is to incorporate joint
learning and control policy synthesis, for instance, PAC-MDP methods
\cite{Fu-RSS-14}, into multi-objective \ac{mdp}s with temporal logic
constraints.

Another limitation in modeling is that the current cognitive model
cannot capture all possible influences of human's cognition on his
performance. Consider, for instance, when the operator is bored or
tired, his performance in some tasks can be degraded, and therefore
the transition probabilities in $M_H$ are dependent on the operator's
cognitive states. In this case, we will need to develop a different
product operation for combining the three factors: $M_A$, a set of
$M_H$'s for different cognitive states, and $M_\att$, into the shared
autonomy system.  Despite the change in modeling the shared autonomy
system, the method for solving Pareto optimal policies developed in
this paper can be easily extended.

\appendix
\section{Weight normalization for multi-objective criteria}
\label{sec:normalization}
Consider a multiobjective \ac{mdp} $\calM=\langle V, \Sigma, \Delta,
\bm D_0, \vec{r},\gamma \rangle$ where $\vec{r}= (r_1,r_2,\ldots,r_n)$
is a vector of reward functions and $\gamma$ is the discount
factor, let $U_{i}(\cdot, f_i^\ast)$ be the vectorial value function
optimal for the $i$-th criterion, specified with the reward function
$r_i$. An approximation of the Nadir point for the $i$-th criterion is
computed as follows, $U_i^N=\sum_{v\in V} \bm
D_0(v)\min_{j=1,\ldots,n}U_{i}(v, f_j^\ast)$ where
$U_i(\cdot,f_j^\ast)$ is a vector value function obtained by
evaluating the optimal policy for the $j$-th criterion with respect to
the $i$-th reward function. The weight vector after normalization is
defined as
$
\lambda_i=\frac{w_i}{\abs{U_i^I-U_i^N}} \enspace.
$

\bibliographystyle{IEEEtran}

\end{document}